\def\year{2019}\relax
\documentclass[letterpaper]{article} 
\usepackage{aaai19}  
\usepackage{times}  
\usepackage{helvet}  
\usepackage{courier}  
\usepackage{url}  
\usepackage{graphicx}  

\usepackage{amssymb, amsmath}
\usepackage{graphicx}
\usepackage{todonotes}
\usepackage[ruled,linesnumbered,algo2e]{algorithm2e}
\usepackage{mathrsfs}
\usepackage{booktabs}
\usepackage{subfigure}
\usepackage{tikz}
\usepackage{pgfplots}

\newcommand{\xx}{\mathbf{x}}
\newcommand{\zz}{\mathbf{z}}

\newcommand{\ww}{\mathbf{w}}
\newcommand{\vv}{\mathbf{v}}
\newcommand{\pp}{\mathbf{p}}
\newcommand{\qq}{\mathbf{q}}
\newcommand{\MM}{\mathbf{M}}

\newcommand{\cX}{\mathcal{X}}
\newcommand{\cY}{\mathcal{Y}}

\newcommand{\sS}{\mathscr{S}}
\newcommand{\Real}{\mathbb{R}}

\newcommand{\aaa}{\boldsymbol{\alpha}}
\newcommand{\mmm}{\boldsymbol{\mu}}
\renewcommand{\top}{\intercal}

\makeatletter
\def\overUnderArrow{\@ifnextchar[\overUnderArrow@i{\overUnderArrow@i[]}}
\def\overUnderArrow@i[#1]#2#3{
  \ifx\relax#1\relax\array[b]{c}\overset{\text{#2}}{\downarrow}\\#3\endarray
  \else\ifx\relax#2\relax
    \array[t]{c}#3\\\underset{\text{#1}}{\uparrow}\endarray
  \else
    \array{c}\overset{\text{#2}}{\uparrow}\\#3\\\underset{\text{#1}}{\downarrow}\endarray
  \fi\fi}
\makeatother

\SetKwComment{Comment}{$\triangleright$\ }{}

\begin{document}
%
\title{Interpretable preference learning: a game theoretic framework for large margin on-line feature and rule learning}
\author{Mirko Polato\ and Fabio Aiolli\\
University of Padova\\
Department of Mathematics\\ 
Via Trieste, 63, 35121 Padova - Italy
}


\maketitle


\begin{abstract}
A large body of research is currently investigating on the connection between machine learning and game theory. In this work, game theory notions are injected into a preference learning framework. 
Specifically, a preference learning problem is seen as a two-players zero-sum game. An algorithm is proposed to incrementally include new useful features into the hypothesis. This can be particularly important when dealing with a very large number of potential features like, for instance, in relational learning and rule extraction.
A game theoretical analysis is used to demonstrate the convergence of the algorithm. Furthermore,
leveraging on the natural analogy between features and rules, the resulting models can be easily interpreted by humans. 
An extensive set of experiments on classification tasks shows the effectiveness of the proposed method in terms of interpretability and feature selection quality, with accuracy at the state-of-the-art.
\end{abstract}

\section{Introduction}

Nowadays, the connection between game theory (GT) and machine learning (ML) is heavily studied in the field of ML for security in which security challenges are faced using integration frameworks between GT and ML~\cite{Yufei:2017,Xu:2017}. 
Many of these approaches can also be considered part of the more general literature concering Adversarial Learning~\cite{Good:2014} that is now a hot topic in ML.
Less recently, the GT-ML duo has been also investigated in the mainstream ML literature. For instance, it is well known that hard margin SVM can be cast into a two-players zero-sum game~\cite{Aiolli:2008}. Another famous example is the seminal work that introduced Adaboost~\cite{Freund:1997}, in which GT is applied to on-line learning. Many others ML concepts have been also related to GT, e.g., as showed by 
Ionnadis et al.~\cite{Ioannidis:2013}, linear regression problems can be seen as a non-cooperative game, while in~\cite{Rezek:2008}, authors elucidate the equivalence between inference in ML and GT.

In this paper we present a principled algorithm inspired by GT and preference learning~\cite{Frnkranz:2010} for classification. The learning problem is seen as a two-players zero-sum game, in which the considered hypotheses spaces consist in a set of preference prototypes along with (possibly non-linear) features. Moreover, we show how feature selection naturally comes as a side effect of the algorithm. 
One of the biggest challenges in feature selection is dealing with large scale data in particular with (infinitely) many input features. This is a typical scenario in real-world applications when data instances are of high dimensionality or it is expensive/inconvenient to acquire all attributes.
In these contexts, batch approaches are simply not applicable for computational reasons. Thus, there is the need to move towards online feature selection (OFS) approaches~\cite{Wang:2014,Wu:2013}, which are allowed to work with a small and limited number of features.
Following this direction, we discuss how the on-line feature generation plays a key role in the algorithm, especially when it comes to interpret the solution of the model, which is very useful for producing explanations.
There are plenty of applications in which explanation plays a key role, such as bioinformatic applications, recommender systems, and support systems for physicians, just to mention a few. 
The notion of explanation is also one of the most controversial subject of the recently introduced GDPR (General Data Protection Regulation). 

To summarize, the main contributions of this work are the following:
\begin{itemize}
	\item a new large margin preference learning method based on game theoretical concepts that naturally offers feature and rule selection capability. The proposed framework is general enough to deal with different kinds of features and rules that are very useful when interpretability is desired;
	\item an efficient approximated algorithm for large scale game matrices with potentially infinite number of columns, allowing the application of the method to online scenarios (OFS). Empirical evidence shows that the number of selected columns by the algorithm is typically orders of magnitude less than the total number of available columns;
	\item an extensive set of experiments is reported to assess both the effectiveness and the quality of the feature/rule selection of our method. Results demonstrate that the algorithm is able to provide sparse solutions that are suitable when the explanation of the decision is required/desirable.
\end{itemize}

The reminder of this paper is structured as follows: next section will introduce all the background knowledge needed to fully understand the notions discussed afterwards. Then, the main contributions of this paper will be described. Finally, all the performed experiments are reported and thoroughly discussed.
For extensive details about the experiments and the datasets please refer to the supplementary material.

\section{Background}
\subsection{Preference Learning}
Preference learning (PL) is a machine learning subfield in which the goal is to learn a preference model from observed item preferences. The constructed model has to predict preferences for previously unseen items.
Label ranking is one of the three main PL tasks~\cite{Frnkranz:2010}: given a set of input patterns $\xx_i \in \cX$, $i \in [1,\dots,n]$, and a finite set of labels $\cY \equiv \{y_1, y_2, \dots, y_m\}$ the goal is to learn a scoring function $g_\theta : \cX \times \cY \rightarrow \Real$ which assigns a score for each instance-label pair $(\xx, y)$.
Label ranking represents a generalization of a classification task, since $g_\theta$ implicitly defines, for an instance $\xx$, a total order over $\cY$.
In the context of PL, the training set used to build the model consists of a set of pairwise preferences $y_i \succ_\xx y_j$, $i\neq j$, i.e., for the pattern $\xx$, $y_i$ is preferred to $y_j$. In the case of classification, in which a pattern $\xx$ is associated to a unique label $y_i$, the set of preferences is $\{y_i \succ_\xx y_j \mid 1 \leq j \neq i \leq m\}$.

In this work we focus on linear preference models~\cite{Tsochantaridis:2004,Aiolli:2010} of the form $g_\theta(\xx, y) = \ww^\top \psi(\xx, y)$, where $\theta \equiv \ww$ is the parameters vector and $\psi : \cX \times \cY \rightarrow \Real^{d \cdot m}$, $\cX \equiv \Real^{d}$, $\cY \equiv \{1,\dots,m\}$ is a joint representation of instance-label pairs.
In order to properly rank the labels w.r.t. each item, given a preference $y_i \succ_\xx y_j$ then $g_\theta(\xx, y_i) > g_\theta(\xx, y_j)$ should hold, and thus
$$
	\ww^\top \psi(\xx, y_i) > \ww^\top \psi(\xx, y_j) \Rightarrow  \ww^\top (\psi(\xx, y_i) - \psi(\xx, y_j)) > 0,
$$
which can be interpreted as the margin (a.k.a. \textit{confidence}) of the preference. Intuitively, large margins correspond to good generalization capability of the ranker~\cite{Schapire:1997}.
We assume an instance-label joint representation defined as $\psi(\xx, y) = \xx \otimes \mathbf{e}^m_y$, where the symbol $\otimes$ indicates the Kronecker product 
and $\mathbf{e}^m_y$ is the $y$-\emph{th} vector of the canonical basis of $\Real^m$.
Thus, given a preference $y_i \succ_\xx y_j$ we construct its corresponding representation by $\zz = \psi(\xx, y_i) - \psi(\xx, y_j) = \xx \otimes (\mathbf{e}^m_{y_i} - \mathbf{e}^m_{y_j})$, $\zz \in \Real^{d \cdot m}$. 
The $f$-th $m$-dimensional chunk of a preference $\zz$ is indicated by
$$
	\zz[f] = (z_{(f-1)\cdot m}, z_{(f-1)\cdot m+1}, \dots, z_{f\cdot m-1}) \in \Real^m.
$$


At classification time, given a new pattern $\xx$, the predicted label $\hat{y}$ is computed by selecting the label that maximizes the score $g_\theta(\xx,y)$, that is, $\hat{y} = \arg\max_{y \in \cY} g_\theta(\xx,y)$.

\subsection{Game Theory}
Game theory is a branch of mathematics that studies the strategic interaction between rational decision-makers.
For the purposes of this paper, we focus on two-players zero-sum games, which are by definition non-cooperative games. The strategic form of a two-players zero-sum game is defined by a matrix $\MM$, dubbed payoff matrix or game matrix. 
The game takes place in rounds in which the two-players, the row player \texttt{P} and the column player \texttt{Q}, play simultaneously: the row player picks a row and the column player picks a column of $\MM \in \Real^{P \times Q}$, where $P$ and $Q$ are the number of available strategies for \texttt{P} and \texttt{Q} respectively. 
Each matrix entry $\MM_{i,j}$ represents the loss of \texttt{P}, or equivalently the payoff of \texttt{Q}, when the strategies $i$ and $j$ are played by the two-players.  
The goal of the player \texttt{P} is to define a strategy that minimizes its expected loss $V$. Conversely, the player \texttt{Q} aims at finding a strategy that maximizes its payoff.
Typically, the players strategies are randomized over the rows/columns of the game matrix: player \texttt{P} selects a row according to a probability distribution $\pp$ over the rows, and, similarly, player \texttt{Q} selects a column according to a probability distribution $\qq$ over the columns.
These strategies are usually referred to as mixed strategies. For presentation purposes we refer to the vectors $\pp$ and $\qq$ as stochastic vectors, that is $\pp \in \sS_{P}$ and $\qq \in \sS_{Q}$, where $\sS_{P} = \{\pp \in \Real_{+}^{P} \mid \|\pp\|_1 = 1\}$ and $\sS_{Q} = \{\qq \in \Real_{+}^{Q} \mid \|\qq\|_1 = 1\}$.
The optimal pair of strategies $(\pp^*,\qq^*)$, i.e., the saddle-point (a.k.a. Nash equilibrium) of $\MM$, has a well-know formulation~\cite{vonNeumann:1928:TGG}, that is
$$
	V^* = {\pp^*}^{\top}\MM \qq^* = \min_{\pp} \max_{\qq} \pp^{\top} \MM \qq = \max_{\qq} \min_{\pp} \pp^{\top} \MM \qq,
$$
where $V^*$ is the value of the game.
It is well known that the saddle-point solution of the equation above can be found using linear programming with a number of variables proportional to the number of (pure) strategies.
It is evident that for high dimensional game matrices the computational complexity can become prohibitive.
A way to overcome this computational issue is to rely on approximated solutions. 
An adaptive approach to compute an approximate saddle-point strategy using multiplicative weights has been proposed by Freund et al. \cite{freundschapire99,DBLP:conf/colt/FreundS96}.
This algorithm, called \emph{Adaptive multiplicative weights} (AMW), is guaranteed to come close to the minimum loss achievable by any fixed strategy. 
More recently, an incremental version of AMW (i-AMW) has also been proposed in~\cite{Bopardikar:2014}. A randomized approach is presented in~\cite{Bopardikar:2010}, where each player chooses its best mixed strategy on a sampled set of rows/columns, that is, the game matrix is a submatrix of the whole payoff matrix. Authors prove that when the submatrix is sufficiently large the achieved approximation is good with high probability.

The \emph{fictitious play} algorithm~\cite{brown51} (a.k.a. Brown-Robinson learning process) is one of the first methods for computing approximate saddle-point strategies. 
Fictitious play starts with a random initial pure strategy for the player \texttt{P}. Then, in turn, each player picks its next pure strategy as the best response, assuming the opponent 
picks uniformly at random one among its previous choices. In other words, at each round both players try to infer the opponent mixed strategy from its previous picks. The pseudo-code of FictPlay adapted to our purposes is reported in Alg.~\ref{alg:fp}.

\begin{algorithm2e}
    \caption{FictPlay: Fictitious Play algorithm}
    \label{alg:fp}
	\DontPrintSemicolon
	\KwIn{
		$\quad \MM \in \Real^{P \times Q}$: matrix game,\\
		\hspace{3.35em}$\quad T_e$: number of iterations
	}
	\KwOut{
		$\quad \pp, \qq$: row/column player strategy,\\
		\hspace{3.9em}$\quad V$: the value of the game
	}
	\BlankLine
	$r \gets randint[1,P]$ \;
	$\mathbf{s}_p, \vv_p \gets \mathbf{0}, \mathbf{0}$ \; 
	$\mathbf{s}_q, \vv_q \gets \MM_{r,:}, \mathbf{e}^{P}_r$ \;
	\BlankLine
	\For {$t \gets 1 \ \mbox{\bf to}\ T_e$} {
		$\hat{q} \gets \arg\max \mathbf{s}_q$, ~~$\mathbf{s}_p \gets \mathbf{s}_p + \MM_{:,\hat{q}}$  \; 
		$\hat{p} \gets \arg\min \mathbf{s}_p$, ~~$\mathbf{s}_q \gets \mathbf{s}_q + \MM_{\hat{p},:}$ \;
		$\vv_q \gets \vv_q + \mathbf{e}_{\hat{q}}^{Q}$, ~~~$\vv_p \gets \vv_p + \mathbf{e}_{\hat{p}}^P$
    }
    $\pp \gets \vv_p / \|\vv_p\|_1$, $\qq \gets \vv_q / \|\vv_q\|_1$, $V \gets \pp^\top \MM \qq$ \;
    \KwRet{$\pp, \qq, V$}
\end{algorithm2e}
In the algorithm, 
$\MM_{r,:}$ and $\MM_{:,c}$ indicate the $r$-th row and the $c$-th column of the matrix $\MM$, respectively. 

\section{A game theoretic perspective of PL}
In this section we introduce the main theoretical contribution of the paper. Specifically, we propose a new learning approach for label ranking based on game theory. We assume to have a set of training preferences of the form $(y_+ \succ_{\xx} y_-)$ which can be easily transformed to their corresponding vectorial representation as described above.
We consider an hypothesis space $\mathcal{H}$ composed by linear functions, i.e., $\mathcal{H} \equiv \{\zz \mapsto \ww^\top \zz \mid \ww,\zz \in \Real^{d \cdot m} \}$.
We say that a preference $\zz$ is satisfied w.r.t. an hypothesis $\ww$ iff $\ww^\top \zz > 0$, that is, whether the margin of the preference $\rho(\zz) = \ww^\top \zz$ is strictly positive. Such margin represents the confidence of the hypothesis $\ww$ over the preference $\zz$. 
According to the maximum margin principle, we aim to select $\ww$ such that it maximizes the minimum margin over the training preferences.
Thanks to the Representer Theorem~\cite{Kimeldorf:1971,Hofmann:08} we know that the optimal $\ww$ can be defined as a convex combination of the training preferences, that is $\ww \propto \sum_j \alpha_j \zz_j, \aaa \in \sS_{P}$.
Thus, we can rewrite the margin of a preference $\zz$ as
\begin{align*}
\rho(\zz) &= \sum_j \alpha_j \zz_j^\top \zz = \sum_j \alpha_j \sum_f \mu_f \zz_j[f]^\top \zz[f] \\
		  &= \sum_j \sum_f \alpha_j\mu_f \zz_j[f]^\top \zz[f] = \sum_{(j,f)} q_{(j,f)} \zz_j[f]^\top \zz[f],
\end{align*}
where the dot product $\zz_j^\top \zz$ is generalized by giving different weights to the features according to a distribution $\mmm$ over the features, and $\qq$ such that $q_{(j,f)} = \alpha_j \mu_f$ is a new distribution over all the possible preference-feature pairs.

Now, let us call $\pp$ the distribution over the training preferences, then the expected preference margin w.r.t. the distribution $\pp$ is defined by
\begin{equation}
\bar{\rho}(\pp,\qq) = \sum_i p_i \sum_{(j,f)} q_{(j,f)} \zz_i[f]^\top \zz_j[f] = \pp^{\top} \MM \qq
\label{eq:margin_pq}
\end{equation}
where $\MM_{i,(j,f)} = \zz_i[f]^{\top} \zz_j[f]$.

If we take a closer look at Eq.~\eqref{eq:margin_pq}, we can easily grasp the strong relation between such a preference learning problem and the game theory setting.
In particular, let us consider a two-players zero-sum game where the row player \texttt{P} (the nature) picks a distribution over the whole set of training preferences (i.e., rows) aiming at minimizing the expected margin. Simultaneously, the opponent player \texttt{Q} (the learner) picks a distribution over the set of preference-feature pairs (i.e., columns) aiming at maximizing the expected margin (payoff). 
So, the value of the game, that is the maximal minimum margin solution is given by
$$V = \bar{\rho}(\pp^*,\qq^*) = \min_{\pp} \max_{\qq} \pp^{\top} \MM \qq$$ 

Thus, it turns out that the distribution $\qq$ maximizing the minimum margin in the training set can be found as the saddle-point solution of the game matrix $\MM$.

\section{Approximating the optimal strategies}
Dimensionally speaking the game matrix $\MM$ in Eq.~\eqref{eq:margin_pq} can be huge, in particular, this is the case for its number of columns as it is equal to the number of all possible preference-feature pairs. Thus, solving the game using standard off-the-shelf algorithms from game theory is impractical.
In this section we propose a new method to overcome this issue by solving the game incrementally. 

The main idea is to iteratively consider only a subset of columns of the whole game matrix, in such a way that, at each iteration, the suboptimal computed solution becomes closer and closer to the optimal one.
Formally, let us suppose to have a game matrix $\MM$ and let denote with $(\pp^*,\qq^*, V^*)$ its corresponding optimal solution. At each iteration we consider a subset of columns of $\MM$, that is $\MM_t = \MM \mathbf{\Pi}_t$ where $\mathbf{\Pi}_t \in \{0,1\}^{Q\times B}$ are left-stochastic (0,1)-matrices, i.e. matrices whose entries belong to the set $\{0,1\}$ and whose columns add up to one. $B$ is the number of columns considered at each iteration, $B \ll P$.
Let $(\pp_t^*,\qq_t^*, V_t^*)$ be the solution for the matrix $\MM_t$ computed at iteration $t$. 
At the end of each iteration, the columns of $\MM_t$ corresponding to null entries in $\qq_t^*$ are replaced by new columns randomly drawn from the whole set of columns.  
We can show that the value of the game at each iteration increases monotonically and it is upper bounded by the optimal margin, that is the value of the game when considering the full matrix $\MM$.
Specifically, let us assume to be at iteration $t+1$: a new left-stochastic (0,1)-matrix $\mathbf{\Pi}_{t+1}$ is considered which is  $\mathbf{\Pi}_{t}$ where every column corresponding to null entries in $\qq_t^*$ are substituted with a new random stochastic vector $\mathbf{e}_{h}^Q$ for a random column $h$. Thus, it can be shown that 
\begin{align} 
\label{dim:1}
V_t^*   = \pp_t^{*\top} \MM_t \qq_t^* & = \pp_{t}^{*\top} \MM \mathbf{\Pi}_t \qq_t^*\\ 
\label{dim:2}
&\leq \pp_{t+1}^{*\top} \MM \mathbf{\Pi}_t \qq_t^* \\ 
\label{dim:3}
& = \pp_{t+1}^{*\top} \MM \mathbf{\Pi}_{t+1} \qq_t^* \\ 
\label{dim:4}
& \leq \pp_{t+1}^{*\top} \MM_{t+1} \qq_{t+1}^* = V_{t+1}^*
\end{align}
and $$\forall t, V_t^* = \pp_{t}^{*\top} \MM \mathbf{\Pi}_t \qq_t^* \leq \pp^{*\top} \MM \underbrace{\mathbf{\Pi}_t \qq_t^*}_{\hat{\qq}_t} \leq \pp^{*\top} \MM \qq^* = V^*.$$

Equivalence~\eqref{dim:1} is trivial since $\MM_t = \MM \mathbf{\Pi}_t$ by definition. Inequality~\eqref{dim:2} holds because the strategy $\pp_{t+1}^*$ is suboptimal for $\MM_t$. 	In~\eqref{dim:3} we simply replace columns of the game matrix corresponding to null entry of $\qq_{t}^*$ which does not affect the value. Finally, inequality~\eqref{dim:4} is true because $\qq_{t}^*$ is suboptimal for $\MM_{t+1}$, and similar considerations can be done for the last series of inequalities.
The pseudo-code of the full algorithm (PRL: Preference and Rule Learning algorithm) is given in Alg.~\ref{alg:our}.


\begin{algorithm2e}[h]
    \caption{PRL: Preference and Rule Learning}
    \label{alg:our}
	\DontPrintSemicolon
	\KwIn{$\mathcal{P}$: set of training preferences\\
		$\quad\quad\quad F_\textit{gen}:$ random feature generator\\
		$\quad\quad\quad B$: size of the working set\\ 
		$\quad\quad\quad T$: number of epochs \\
		$\quad\quad\quad T_e$: number of iterations of \textit{FictPlay}
	}
	\KwOut{$\mathcal{Q}$: working set of hypothesis\\
		$\quad\quad\quad\;\;\; \qq$: mixed strategy in $\mathcal{Q}$\\ 
	}
	\BlankLine
	random initialization of the set $\mathcal{Q}$ such that $|\mathcal{Q}|=B$ \;
	compute the matrix game $\MM$ on the basis of $\mathcal{P}$ (rows) and $\mathcal{Q}$ (cols) \;
	\For{$t \gets 1 \ \mbox{\bf to}\  T$}{
		$\pp, \qq, v \gets \textit{FictPlay}(\MM,T_e)$ \;
		\If{$t < T$}{
			\ForEach{$(j,f) \mid \qq_{(j,f)} = 0$}{
				$(j', f') \gets \textit{pick}(\mathcal{P}), F_\textit{gen}()$ \;
				update $\mathcal{Q}$: replace $(j,f)$ with $(j',f')$ \;
				update columns of $\MM$ w.r.t. $\mathcal{Q}$: \;
				\hspace{1cm}let $k$ the position of $(j',f')$ in $\mathcal{Q}$,\\ 
				\hspace{1cm}for all $i \in \mathcal{P}$, $\MM_{i,k} = \zz_i[f]^{\top}\zz_{j'}[f]$
				
			}
		}
    }
    \KwRet{$\qq, \mathcal{Q}$}
\end{algorithm2e}

It is worth to notice that the algorithm does not require any prior knowledge about $\MM$, and the number of columns can be also infinite. Thus, a natural approach for dealing with potentially infinite game matrices is to use an online column generation approach.

\subsection{Online feature/rule generation}
One of the most important steps of the PRL is the generation of new columns. Since a game matrix column is defined by a preference-feature pair we need to define how such features are generated. In this work we focus on two features generation schemes: polynomial feature generation, and rule generation.
In the algorithm we referred to a generic feature generator scheme with the function $F_\textit{gen}$.

\subsubsection{Polynomial features generation}
This scheme generates features that are taken from the space of features of a polynomial kernel. In particular, we focus on homogeneous polynomial features of a given degree. For example, given an $n$-dimensional instance $\xx$ some possible polynomial features of degree 3 are: $x_1x_2x_n$, $x_1^2x_3$ and $x_n^3$. Note that, when the input variables are binary-valued such monomials correspond to logical conjunctions.

\subsubsection{Rules generation}
A rule is a logical condition over an input variable. The introduction of rules can be useful when it comes to interpret the model.  
Let us make some examples. In the case of binary valued input variables a rule is simply its truth value. When dealing with continuous variables a rule is a relation involving the values of the variables. For instance, by defining a threshold like $x \geq 5$, or equalities such as $x = 3.2$.
Such thresholds can be chosen on the basis of an heuristic or randomly from the set of unique values of the variable.
In order to create more complex rules, it is also possible to combine them. In particular, given two rules, their product represents, from a logical stand point, the conjunction of the two conditions. In the remainder we will refer to the arity of this combination as the degree of the rule. 

\section{Evaluation}
In this section we describe and discuss the set of experiments performed to assess the effectiveness of PRL. 
Specifically, we performed two different sets of experiments: the first set aims to evaluate the degree of interpretability of PRL, while the second set focuses on the assessment of the performance and the quality of the feature selection. In all the experiments the number of iterations $T_e$ of FictPlay has been set to $10^6$, while the number $T$ of epochs of PRL has been set to $10^3$.
The complete set of experiments as well as a thorough analysis of the obtained results is reported in the supplementary material. In the following sections we present the most relevant and interesting results.

\subsection{Model interpretation}
In the first set of experiments, we employed PRL to select the most relevant features for interpreting the model. The aim is to use these features to explain the decision. 
We run PRL on four benchmark datasets:
\begin{description}
	\item[\texttt{tic-tac-toe}] is a dataset containing 958 ending positions of the game tic-tac-toe, and the task is to classify whether the $\times$ is the winner;
	\item[\texttt{breast-cancer}] is the well known Breast Cancer Wisconsin Diagnostic Dataset, where the task is to classify a tumor as malignant or benign. For more details about the dataset please refer to~\cite{Hayashi:2015};
	\item[\texttt{poker}] dataset contains 25010 poker hands and the task is to classify the value of the hand, e.g., pair, full house and so on (10 classes). In our experiments, three binary classification tasks have been derived from the original multiclass dataset as described in the following section;
	\item[\texttt{mnist}] is a (well known) dataset of handwritten digits. The task consists in classifying the digits (10 classes).
 \end{description}

The datasets have been pre-processed as in the following: \texttt{tic-tac-toe} has been converted into a binary-valued dataset through one-hot encoding, obtaining 27 binary input variables for each instance. Both \texttt{breast-cancer} and \texttt{mnist} did not require any pre-processing. For every datasets, instances with missing values have been removed.
For the \texttt{poker} dataset we defined a hierarchy of features. The goal was showing that, increasing the expressiveness of the features, PRL is still able to identify the smallest set of features/rules able to explain the classification. The hierarchy can be summarized as follows:
\begin{description}
	\item[Level 1] The features are a simple enumeration of the cards. An hand is represented with a binary vector of dimension 52 in which an entry is equal to 1 iff the corresponding card is in the hand;
	\item[Level 2] Features which are a counting aggregation of the suits and the ranks of the cards in the hand are added. Suits are represented as a four dimensional vector, while ranks as a 13 dimensional vector. Note that the Ace is assumed of rank 1, while J=11, Q=12, and K=13;
	\item[Level 3] This level contains 5 additional features that are a further aggregation of the features in the first and second level, namely: number of different ranks, number of different suits, number of cards of the most popular suit, number of cards of the most popular rank, and maximum difference between ranks.
\end{description}
Note that, at each level of the hierarchy the features of the previous level are kept in the representation. For extensive details about the construction of the hierarchy please refer to the supplementary material. 

\subsubsection{The \texttt{tic-tac-toe} dataset}
The tic-tac-toe dataset has been used as a toy testbed in which the positive class (i.e., win for $\times$) can be expressed with a single DNF rule. This experiment aims to show that PRL is able to identify such explaining rules.
Since we have a-priori knowledge about the game, we trained PRL using polynomial features of degree 3. In the case of binary valued data, polynomial features correspond to conjunctions, and hence they are suited for our purposes.
Experiments have been performed using a 70-30\% training and test split division.
After the training the 10 features with the highest weights were: $x_{8}x_{17}x_{26}$, $x_{2}x_{11}x_{20}$, $x_{2}x_{14}x_{26}$, $x_{8}x_{14}x_{20}$, $x_{20}x_{23}x_{26}$, $x_{11}x_{14}x_{17}$, $x_{2}x_{5}x_{8}$, $x_{5}x_{14}x_{23}$, $\tilde{x}_{13}^3$ and $\tilde{x}_{25}^3$. 
Features with a $\sim$ on top are the ones which characterize a negative preference (no win for $\times$). The main observation about these features is that the first 8 are in fact the available three-in-a-row for the crosses, while the last two features represent a naught in the central and in the bottom right cell. The former is reasonable since it is more likely to win by occupying the central square, the latter is instead not very significant. Overall, the algorithm has been able to identify all the conditions that determine a win for the cross.

\subsubsection{The \texttt{poker} dataset}
Akin to the \texttt{tic-tac-toe}, in the \texttt{poker} dataset the classes can be described by means of logical rules. However, in this case the rules are generally much more complex. As described earlier, a three levels hierarchy of features has been defined, with the aim of testing whether the algorithm is able to recognize the smallest set of useful features to explain the classification.
The main idea is that the algorithm should be able to retrieve the easiest rules which explain the classification when more expressive features are also available.

In order to have more control on the tasks, and to highlight the different difficulty in recognizing hand values such as a straight, we defined the following three binary classification tasks: \textit{TOK} (Three Of a Kind) versus rest, \textit{Flush} versus rest, and \textit{Straight} versus rest.
Besides the rule extraction test, we also compared the balanced accuracy (BACC) of PRL w.r.t. a standard SVM with polynomial kernel. 
We employed the BACC because the dataset is highly imbalanced, i.e., positive class $\leq 2\%$. The balanced accuracy is defined as:
$$
	\text{BACC} = \frac{1}{2}\left(\frac{TP}{P} + \frac{TN}{N}\right) \times 100,
$$
where $TP$ stands for true positives ($P$ = positives), and $TN$ for true negatives ($N$ = negatives).

The hyper-parameter $C$ of SVM has been validated in the set $\{10^{-3}, \dots, 10^4\}$, and the degree in the range [1,3] via 3-fold cross validation. PRL has been trained using rules of degree 1 on the set of relations $\{=\}$. Experiments have been performed using a 80-20\% training and test split division.

The achieved results are reported in Table~\ref{tab:poker-svm}.

\begin{table}[h!]
\centering
\setlength{\tabcolsep}{.7em}
\begin{tabular}{lcccc}
	\toprule
	\textbf{Method} & \textbf{\# Level} & \textbf{TOK} & \textbf{Flush} & \textbf{Straight} \\
	\toprule
	SVM & 1 & \textbf{50.00} & 50.00 & \textbf{50.00} \\
	PRL & 1 & 48.08 & \textbf{59.03} & 49.97 \\ \midrule
	SVM & 2 & 50.00 & 50.00 & 50.00 \\
	PRL & 2 & \textbf{100.00} & \textbf{77.75} & \textbf{51.29}\\ \midrule
	SVM & 3 & 99.99 &  96.43 & 50.00\\
	PRL & 3 & \textbf{100.00} & \textbf{100.00} & \textbf{100.00} \\
	\bottomrule
\end{tabular}
\caption{Balanced accuracy (\%) on the \texttt{poker} dataset. The highest accuracies in all classification tasks, and in all levels, are highlighted in \textbf{bold}. \label{tab:poker-svm}}
\end{table}

The first observation worth to be mentioned is that PRL using the third level of features has been able to identify, for all the tasks, the rules which explain the classification, achieving perfect accuracy. 
A remarkable difference w.r.t. the SVM can be noticed in the \textit{Straight} classification, which is the hardest task. SVM simply classifies any instance as negative ($TP = 0$), with the same behaviour in all the tasks both at the second and at the first level of the hierarchy. Concerning the third level, SVM had one false negative in the \textit{TOK} task and one false positive in the \textit{Flush}, achieving a BACC of 99.99\% and 96.43\%, respectively.

Let us now analyze the rules extracted by PRL in each task. We can observe that none of the tasks can be explained with simple rules in the first level of the hierarchy. In fact, at level 1, the algorithm struggles in all the tasks achieving a BACC around 50\% like SVM, with the exception of the \textit{Flush} in which it achieves a 59\%. 
In the second level, instead, there are features expressive enough to explain both the \textit{TOK} and the \textit{Flush}, while the \textit{Straight} remains a quite hard task. Specifically, an hand contains a TOK anytime one of the rank has a cardinality $= 3$. However, this rule also includes the full house as a false positive. Similarly, a flush can be described with a suit of cardinality $5$, but this also includes the straight flush and the royal flush as false positives. In both these tasks, PRL found the just mentioned rules. It is interesting to note that the method has been able to avoid the full house mistake by adding rules for the negative class (i.e., not a flush). In particular, these rules state that there is a rank with cardinality $=2$, which is enough to say that there is no TOK in the hand.
With the features contained in the third and final level of the hierarchy it is possible to define all the considered classes:
\begin{description}
	\item[\textit{TOK}] \# of ranks = 3, \# of cards of the most popular rank = 3;
	\item[\textit{Straight}] \# of ranks = 5, max difference between ranks = 4, \# of suits $\neq 1$;
	\item[\textit{Flush}] \# of suits = 1, max difference between ranks $\neq$ 4.
\end{description}
At this level, PRL was able to identify all the correct rules achieving a BACC of 100\% in all the tasks.\\

\textbf{Remark}: whenever each class can be defined by simple rules, we showed that PRL is able to identify them all. However, when there are classes which cannot be easily characterized, PRL fails in finding reasonable rules, as for example in the \texttt{poker}. In this case, the positive class (e.g., \textit{Flush}) is defined by simple rules, but there are no easy ways to express its opposite. Thus, PRL will return, with the highest weights, the rules that explain the positive class. All the remaining rules (with small weights) will be associated with negative preferences, but with very small coverage and hence with small generalization capabilities. To address this issue, we allow the feature generation mechanism to pick new pairs consisting of rules which are always true. When such feature is selected and associated with a negative preference  it will give a bias to the negative class. In this way, when the positive rules are not satisfied the decision of the classifier will be the negative class. 
This additional feature has been used in all experiments concerning the \texttt{poker} dataset.

\subsubsection{The \texttt{breast-cancer} dataset}
The Wisconsin Breast Cancer dataset (\texttt{breast-cancer}) is a standard UCI~\cite{Lichman:2013} dataset that contains 682 hospital patients values captured via a Fine-needle aspiration test. Each patient is described by 9 attributes concerning breast tumoral cells. The task consists in classifying a tumor between benign or malignant. The classes distribution is 35\% benign, and 65\% malignant.

Unlike the previous datasets, \texttt{breast-cancer} is a real-world dataset which cannot be explained with simple rules, hence it is not possible to compare the retrieved rules w.r.t. a given ground truth. 
For this reason we compared the rules extracted by PRL with the ones extracted by the rule extraction methods described in~\cite{Hayashi:2015}. 
We directly applied the rules to the entire dataset and the achieved accuracies have been compared. All the methods have been trained using a 90-10\% training and test split. 
This evaluation procedure has been chosen because we only had at our disposal, for each model, the set of extracted rules after a 10-fold cross validation procedure.
We trained PRL using rules of degree 2 on the set of relations $\{\geq, \leq\}$. In Table~\ref{tab:wdbc} the achieved results are summarized.

\begin{table}[h!]
\centering
\setlength{\tabcolsep}{.7em}
\begin{tabular}{lcc}
	\toprule
	\textbf{Method} & \textbf{\# Rules} & \textbf{Accuracy (\%)}\\
	\midrule
	SSV & 3 & 86.36 \\
	GASVM & 2 & 90.03 \\
	C-MLP2LN & 5 & 96.92 \\
	QSVM-G & 12 & 96.48 \\
	ReRXJ48 & 4 & 94.28 \\ 
	\midrule
	PRL only 4th & 1 & 92.67 \\
	PRL@5 & 5 & 96.12 \\
	PRL@10 & 10 & \textbf{97.95} \\
	\bottomrule
\end{tabular}
\caption{Accuracy of the rules extracted by the different algorithms. The highest accuracy is highlighted in \textbf{bold}. \label{tab:wdbc}}
\end{table}

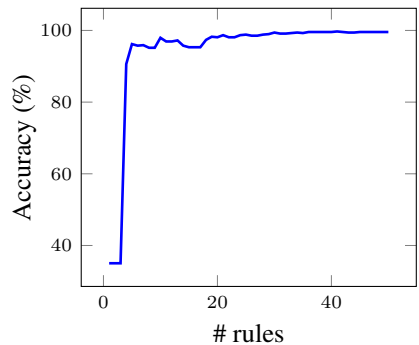
\begin{figure}[h!]
\centering
\begin{tikzpicture}
	\begin{axis}[
			scale=0.65,
			xlabel=\# rules,
			ylabel=Accuracy (\%),
			tick label style={font=\scriptsize},
			ylabel style={yshift=-3ex},
			xlabel style={yshift=1ex},
		]
		\addplot[blue,line width=1pt] table[x=x, y=y] {plot_acc_curve.txt};	
	\end{axis}
\end{tikzpicture}
\caption{Plot of the accuracy w.r.t. the number of considered rules during the classification. \label{fig:curve-acc}}
\end{figure}

As evident from the table PRL achieves the best accuracy using the 10 most relevant rules. Nevertheless, even using fewer rules (i.e., 5) it is able to achieve very good results even though methods such as C-MLP2LN and QSVM-G had slightly higher accuracies. Another interesting observation is that by using the 4th most relevant rule alone the proposed method achieves more than 92\% of accuracy. However, it is also worth to notice that the first three rules are not enough to get good results, as highlighted in Figure~\ref{fig:curve-acc}.

The figure shows the accuracy achieved by considering a limited set of rules (1 up to 50). The algorithm is able to overcome 95\% of accuracy quite rapidly (first four rules) and then it increases almost monotonically until reaching 99.56\% using 50 rules. However, 50 rules are not a reasonable number when it comes to interpret a classification, and thus in Table~\ref{tab:wdbc} we only mentioned the accuracies up to 10 rules.
More details about the extracted rules are reported in the supplementary material.

\subsubsection{The \texttt{mnist} dataset}
The \texttt{mnist} dataset is one of the most widely used dataset for the hand-written digit classification task. The digits are stored in a grey scale 28 by 28 pixel matrix, where each pixel can assume a value between 0 and 255 (0-1 normalized). The task is to recognize the digit represented by an instance. 

\begin{figure}
\centering
\subfigure[0 versus 9]{
\includegraphics[scale=.4]{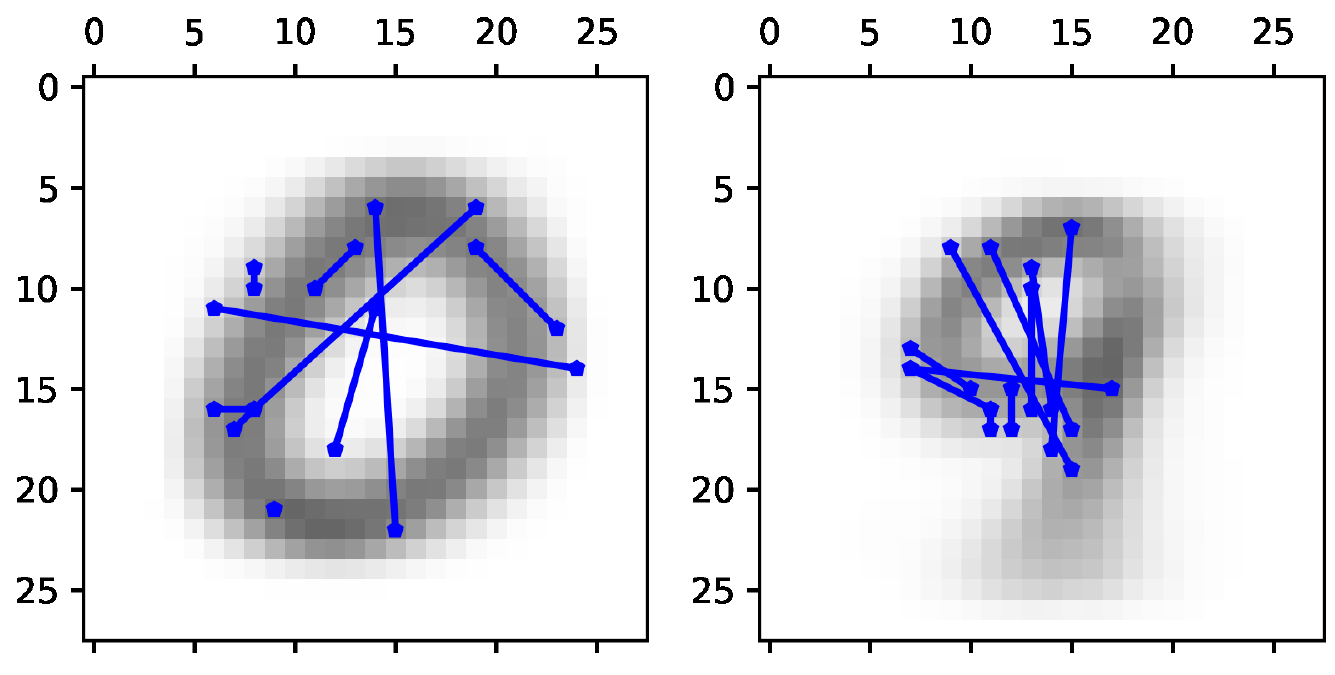}\label{fig:09}
}
\subfigure[4 versus 6]{
\includegraphics[scale=.4]{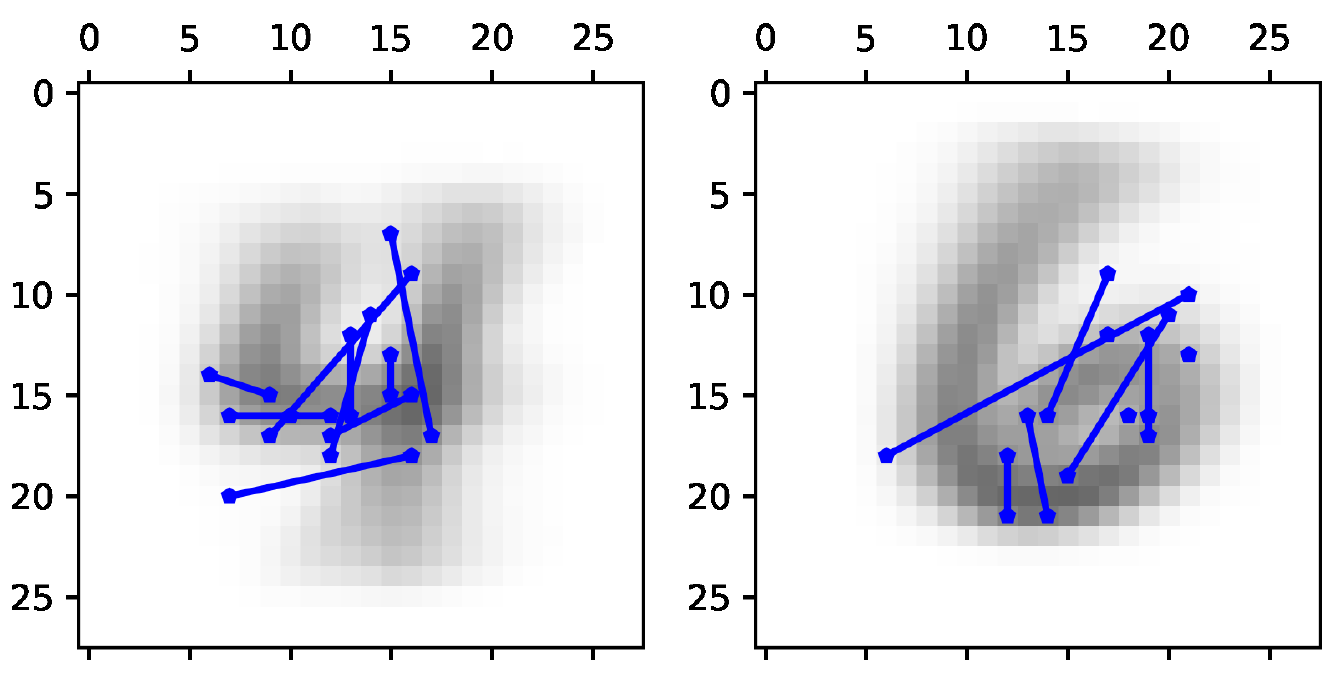}\label{fig:46}
}
\caption{Visualization of the most relevant polynomial features of degree 2. The polynomial features are visualized as segments limited by the involved input variables. The left hand side plot shows the features relevant to discriminate the (a) 0 from the 9 and (b) 4 from the 6. Viceversa in the right hand side plots.\label{fig:mnist}}
\end{figure}

As for \texttt{breast-cancer}, the classification of the \texttt{mnist} digits cannot be explained with simple rules. The goal here is to use the most relevant features for interpreting, in a human fashion, which are the visual characteristics that are leveraged by the model to discriminate each class w.r.t. another.
Experiments have been performed using polynomial features of degree 2. Figure~\ref{fig:mnist} illustrates two examples of the most relevant features used by the model to distinguish: (a) 0 from a 9 (left) and viceversa (right); (b) 4 from a 6 (left) and viceversa (right). 

The features are represented as segments between the two involved variables (i.e., pixels) in each monomial (i.e., rule). In the background the average digit of that class is depicted. 

From the first plots (Figure~\ref{fig:09}) it is clear that in order to discriminate a 0 from a 9 the most important characteristics for the algorithm are the ``big'' curvature for the 0, and the smaller one for the 9. Similarly, in Figure~\ref{fig:46} the same kind of features are important to characterize the 6 w.r.t. the 4, while the 4 is mainly recognized by its horizontal dash.

The complete set of features for every pair of classes have been omitted for space reasons, and can be found in the supplementary material.

\subsection{Feature Selection}
This set of experiments aims to assess the effectiveness of PRL on datasets with many noisy and redundant features. The chosen testbeds have been the datasets of the NIPS 2003 Feature selection challenge \cite{Nips:03}. All the datasets are freely available at the NIPS 2003 Feature selection challenge site, \url{http://clopinet.com/isabelle/Projects/NIPS2003/}.
Further details about the datasets are reported in the supplementary material and in \cite{Nips:03,Johnson:09}. A common characteristic of these datasets is the huge number of features they have compared to the number of training instances. All the datasets consist of binary classification tasks.

We compared PRL with a standard soft-margin SVM. Given the huge number of features of the target datasets, the linear kernel turned out to be a good kernel for these tasks, with the exception of \texttt{madelon} in which the degree 2 polynomial was the best performing kernel for SVM.
The $C$ hyper-parameter of the SVM has been validated in the set of values $\{10^{-4},\dots, 10^5\}$ using a 3-fold cross validation procedure. 
Experiments have been performed using a 70-30\% training and test split. In Table \ref{tab:results-fs} the results achieved by both methods as well as the number of relevant features according to PRL are summarized. In these experiments, the size $B$ of the working set has been set to 2000.
\begin{table}[h!]
\centering
\setlength{\tabcolsep}{.7em}
\begin{tabular}{lccc}
	\toprule
	\textbf{Dataset} & \textbf{SVM} & \textbf{PRL} & \textbf{\# Relevant/Total feat.}\\
	\midrule
	\texttt{dorothea} & 91.88 & \textbf{92.69} & 476/100k\\
	\texttt{gisette} & 96.71 &  \textbf{97.19}& 900/5k\\
	\texttt{madelon} & 60.10 & \textbf{62.75} & 1225/250k\\
	\bottomrule
\end{tabular}
\caption{Accuracy results achieved by SVM and PRL. The last column indicates the number of support preference-feature pairs used by PRL. The best results are highlighted in \textbf{bold}. \label{tab:results-fs}}
\end{table}
As evident from the table, the proposed method is able to achieve better performance than SVM. It is worth to mention that, generally the number of features used by PRL is orders of magnitude less than the number of original features. 

\section{Conclusions and future work}
This paper has proposed a new preference learning framework for classification based on game theoretic concepts. The learning problem is defined as a two-players zero-sum game for which we have given an incremental solution w.r.t. the columns of the game matrix. We provided theoretical guarantees about the convergence of the algorithm as well as an extensive set of experiments demonstrating its effectiveness. We also showed the capabilities of PRL in identifying explanation rules for interpreting the classification.

In this work we only use PRL in classification contexts. In the future we would like to test our method on other PL settings, including instance and label ranking tasks.
From a computational point of view, the algorithm can be further improved starting from the column generation policy. In the future we aim to explore new approaches to select in a smart way the columns which have good chances of being included in the strategy. 
Moreover, we also intend to relax the formulation in order to get a soft margin version of the algorithm. 

Finally, another point of improvement is represented by the random feature generation. A possible future path in this direction can be to explore feature generation methods such as the one proposed in~\cite{Rahimi:2007}. The same algorithm can also be applied to other applications which are based on a large number of explicit features such as classification of structured data (e.g., graphs) or relational learning.

\clearpage
\newpage
\bibliographystyle{aaai}
\bibliography{library}

\begin{thebibliography}{}

\bibitem[\protect\citeauthoryear{Aiolli and Sperduti}{2010}]{Aiolli:2010}
Aiolli, F., and Sperduti, A.
\newblock 2010.
\newblock A preference optimization based unifying framework for supervised
  learning problems.
\newblock In {\em Preference Learning}.

\bibitem[\protect\citeauthoryear{Aiolli, Da~San~Martino, and
  Sperduti}{2008}]{Aiolli:2008}
Aiolli, F.; Da~San~Martino, G.; and Sperduti, A.
\newblock 2008.
\newblock A kernel method for the optimization of the margin distribution.
\newblock In {\em Artificial Neural Networks - ICANN 2008},  305--314.

\bibitem[\protect\citeauthoryear{Bopardikar and
  Langbort}{2014}]{Bopardikar:2014}
Bopardikar, S.~D., and Langbort, C.
\newblock 2014.
\newblock Incremental approximate saddle-point computation in zero-sum matrix
  games.
\newblock In {\em 53rd IEEE Conference on Decision and Control},  1936--1941.

\bibitem[\protect\citeauthoryear{Bopardikar \bgroup et al\mbox.\egroup
  }{2010}]{Bopardikar:2010}
Bopardikar, S.~D.; Borri, A.; Hespanha, J.~P.; Prandini, M.; and Benedetto, M.
  D.~D.
\newblock 2010.
\newblock Randomized sampling for large zero-sum games.
\newblock In {\em 49th IEEE Conference on Decision and Control (CDC)},
  7675--7680.

\bibitem[\protect\citeauthoryear{Brown}{1951}]{brown51}
Brown, G.~W.
\newblock 1951.
\newblock Iterative solutions of games by fictitious play.
\newblock {\em In: Activity Analysis of Production and Allocation}  374--376.

\bibitem[\protect\citeauthoryear{Freund and
  Schapire}{1996}]{DBLP:conf/colt/FreundS96}
Freund, Y., and Schapire, R.~E.
\newblock 1996.
\newblock Game theory, on-line prediction and boosting.
\newblock In {\em COLT},  325--332.

\bibitem[\protect\citeauthoryear{Freund and Schapire}{1997}]{Freund:1997}
Freund, Y., and Schapire, R.~E.
\newblock 1997.
\newblock A decision-theoretic generalization of on-line learning and an
  application to boosting.
\newblock {\em J. Comput. Syst. Sci.} 55(1):119--139.

\bibitem[\protect\citeauthoryear{Freund and Schapire}{1999}]{freundschapire99}
Freund, Y., and Schapire, R.~E.
\newblock 1999.
\newblock Adaptive game playing using multiplicative weights.
\newblock {\em Games and Economic Behavior} 29(1-2):79--103.

\bibitem[\protect\citeauthoryear{F\"urnkranz and
  H\"ullermeier}{2010}]{Frnkranz:2010}
F\"urnkranz, J., and H\"ullermeier, E.
\newblock 2010.
\newblock {\em Preference Learning}.
\newblock Springer, 1st edition.

\bibitem[\protect\citeauthoryear{Goodfellow \bgroup et al\mbox.\egroup
  }{2014}]{Good:2014}
Goodfellow, I.; Pouget-Abadie, J.; Mirza, M.; Xu, B.; Warde-Farley, D.; Ozair,
  S.; Courville, A.; and Bengio, Y.
\newblock 2014.
\newblock Generative adversarial nets.
\newblock In {\em Advances in Neural Information Processing Systems 27}.
\newblock  2672--2680.

\bibitem[\protect\citeauthoryear{Guyon \bgroup et al\mbox.\egroup
  }{2005}]{Nips:03}
Guyon, I.; Gunn, S.; Ben-Hur, A.; and Dror, G.
\newblock 2005.
\newblock Result analysis of the nips 2003 feature selection challenge.
\newblock In Saul, L.~K.; Weiss, Y.; and Bottou, L., eds., {\em Advances in
  Neural Information Processing Systems 17}. MIT Press.
\newblock  545--552.

\bibitem[\protect\citeauthoryear{Hayashi and Nakano}{2015}]{Hayashi:2015}
Hayashi, Y., and Nakano, S.
\newblock 2015.
\newblock Use of a recursive-rule extraction algorithm with j48graft to achieve
  highly accurate and concise rule extraction from a large breast cancer
  dataset.
\newblock {\em Informatics in Medicine Unlocked} 1:9 -- 16.

\bibitem[\protect\citeauthoryear{Hofmann, Scholkopf, and
  Smola}{2008}]{Hofmann:08}
Hofmann, T.; Scholkopf, B.; and Smola, A.~J.
\newblock 2008.
\newblock Kernel methods in machine learning.
\newblock {\em The Annals of Statistics} 36(3):1171--1220.

\bibitem[\protect\citeauthoryear{Ioannidis and Loiseau}{2013}]{Ioannidis:2013}
Ioannidis, S., and Loiseau, P.
\newblock 2013.
\newblock Linear regression as a non-cooperative game.
\newblock In {\em Proceedings of the 9th International Conference on Web and
  Internet Economics - Volume 8289}, WINE 2013,  277--290.

\bibitem[\protect\citeauthoryear{Johnson}{2009}]{Johnson:09}
Johnson, N.
\newblock 2009.
\newblock A study of the nips feature selection challenge.

\bibitem[\protect\citeauthoryear{Kimeldorf and Wahba}{1971}]{Kimeldorf:1971}
Kimeldorf, G.~S., and Wahba, G.
\newblock 1971.
\newblock Some results on tchebycheffian spline functions.
\newblock {\em Journal of Mathematical Analysis and Applications} 33(1):82--95.

\bibitem[\protect\citeauthoryear{Lichman}{2013}]{Lichman:2013}
Lichman, M.
\newblock 2013.
\newblock {UCI} machine learning repository.

\bibitem[\protect\citeauthoryear{Rahimi and Recht}{2007}]{Rahimi:2007}
Rahimi, A., and Recht, B.
\newblock 2007.
\newblock Random features for large-scale kernel machines.
\newblock In {\em Proceedings of the 20th International Conference on Neural
  Information Processing Systems}, NIPS'07,  1177--1184.

\bibitem[\protect\citeauthoryear{Rezek \bgroup et al\mbox.\egroup
  }{2008}]{Rezek:2008}
Rezek, I.; Leslie, D.~S.; Reece, S.; Roberts, S.~J.; Rogers, A.; Dash, R.~K.;
  and Jennings, N.~R.
\newblock 2008.
\newblock On similarities between inference in game theory and machine
  learning.
\newblock {\em J. Artif. Intell. Res.} 33:259--283.

\bibitem[\protect\citeauthoryear{Schapire \bgroup et al\mbox.\egroup
  }{1997}]{Schapire:1997}
Schapire, R.~E.; Freund, Y.; Barlett, P.; and Lee, W.~S.
\newblock 1997.
\newblock Boosting the margin: A new explanation for the effectiveness of
  voting methods.
\newblock In {\em Proceedings of the Fourteenth International Conference on
  Machine Learning}, ICML '97,  322--330.

\bibitem[\protect\citeauthoryear{Tsochantaridis \bgroup et al\mbox.\egroup
  }{2004}]{Tsochantaridis:2004}
Tsochantaridis, I.; Hofmann, T.; Joachims, T.; and Altun, Y.
\newblock 2004.
\newblock Support vector machine learning for interdependent and structured
  output spaces.
\newblock In {\em Proceedings of the Twenty-first International Conference on
  Machine Learning}, ICML '04,  104--.
\newblock New York, NY, USA: ACM.

\bibitem[\protect\citeauthoryear{von Neumann}{1928}]{vonNeumann:1928:TGG}
von Neumann, J.
\newblock 1928.
\newblock Zur theorie der gesellschaftsspiele.
\newblock 100:295--320.

\bibitem[\protect\citeauthoryear{Wang \bgroup et al\mbox.\egroup
  }{2014}]{Wang:2014}
Wang, J.; Zhao, P.; Hoi, S. C.~H.; and Jin, R.
\newblock 2014.
\newblock Online feature selection and its applications.
\newblock {\em IEEE Transactions on Knowledge and Data Engineering}
  26(3):698--710.

\bibitem[\protect\citeauthoryear{Wu \bgroup et al\mbox.\egroup
  }{2013}]{Wu:2013}
Wu, X.; Yu, K.; Ding, W.; Wang, H.; and Zhu, X.
\newblock 2013.
\newblock Online feature selection with streaming features.
\newblock {\em IEEE Transactions on Pattern Analysis and Machine Intelligence}
  35(5):1178--1192.

\bibitem[\protect\citeauthoryear{Xu \bgroup et al\mbox.\egroup
  }{2017}]{Xu:2017}
Xu, H.; Ford, B.~J.; Fang, F.; Dilkina, B.~N.; Plumptre, A.~J.; Tambe, M.;
  Driciru, M.; Wanyama, F.; Rwetsiba, A.; Nsubaga, M.; and Mabonga, J.
\newblock 2017.
\newblock Optimal patrol planning for green security games with black-box
  attackers.
\newblock In {\em GameSec}.

\bibitem[\protect\citeauthoryear{Yufei~Liu}{2017}]{Yufei:2017}
Yufei~Liu, D.~P.
\newblock 2017.
\newblock A novel kernel {SVM} algorithm with game theory for network intrusion
  detection.
\newblock {\em {KSII} Transactions on Internet and Information Systems} 11(8).

\end{thebibliography}

\end{document}